\documentclass[runningheads]{llncs}
\usepackage{graphicx}
\usepackage{amsfonts}
\usepackage{xcolor}
\usepackage{blindtext}
\usepackage{hyperref}
\usepackage{bm}
\begin{document}
\title{Functional2Structural: Cross-Modality Brain Networks Representation Learning}
%
%

\author{Haoteng Tang\inst{1} \and
Xiyao Fu\inst{1} \and
Lei Guo\inst{1} \and
Yalin Wang\inst{2} \and
Scott Mackin \inst{3} \and
Olusola Ajilore \inst{4} \and
Alex Leow \inst{4} \and
Paul Thompson \inst{5} \and
Heng Huang \inst{1} \and
Liang Zhan \inst{1}}
%
%
\institute{University of Pittsburgh, Pittsburgh, PA 15260, USA \and
Arizona State University, Tempe, AZ 85281, USA \and
University of California San Francisco, San Francisco, CA 94143, USA \and
University of Illinois Chicago, Chicago, IL 60607, USA \and
University of Southern California, Los Angeles, CA 90007, USA}

\maketitle              
\begin{abstract}
MRI-based modeling of brain networks has been widely used to understand functional and structural interactions and connections among brain regions, and factors that affect them, such as brain development and disease. 
Graph mining on brain networks may facilitate the discovery of novel biomarkers for clinical phenotypes and neurodegenerative diseases. 
Since brain networks derived from functional and structural MRI describe the brain topology from different perspectives, exploring a representation that combines these cross-modality brain networks is non-trivial. 
Most current studies aim to extract a fused representation of the two types of brain network by projecting the structural network to the functional counterpart. 
Since the functional network is dynamic and the structural network is static, mapping a static object to a dynamic object is suboptimal. 
However, mapping in the opposite direction is not feasible due to the non-negativity requirement of current graph learning techniques. 
Here, we propose a novel graph learning framework, known as Deep Signed Brain Networks (DSBN), with a signed graph encoder that, from an opposite perspective, learns the cross-modality representations by projecting the functional network to the structural counterpart. 
We validate our framework on clinical phenotype and neurodegenerative disease prediction tasks using two independent, publicly available datasets (HCP and OASIS). 
The experimental results clearly demonstrate the advantages of our model compared to several state-of-the-art methods. 

\end{abstract}

\section{Introduction}

Recent years have witnessed great progress in applying graph theory to study MRI-derived  brain  networks, to  discover  novel  biomarkers  for  clinical  phenotypes  or  neurodegenerative  diseases  (e.g.,  Alzheimer’s  disease  or  AD) ~\cite{hao2013multimodal,uludaug2014general,calhoun2016multimodal}. 
Different MRI techniques can be used to reconstruct brain networks corresponding to different aspects of the brain organization or dynamics. 
For example, functional MRI-derived functional networks provide measures of BOLD relationships over time between brain regions but not the existence of a physical direct link between these regions. 
By contrast, the diffusion MRI-derived structural network can describe the connectivity of white matter tracts between brain regions, yet does not inform us about whether this tract, or the regions it connects, are “activated” or “not activated” in a specific state. 
Therefore, these brain networks provide distinct but complementary information, and separately analyzing each of these networks will always be suboptimal.

Graph neural networks (GNNs) \cite{kipf2016semi} have gained enormous attention recently. 
Although some progress~\cite{kawahara2017brainnetcnn,ktena2018metric} has been made by applying GNNs to brain networks, most existing techniques have been developed based on single-modality brain networks. It is well known that a high-level dependency, based on network communications, exists between brain structural and functional networks \cite{rusinek2003regional,bullmore2012economy}, which has motivated many studies \cite{zhang2020deep,huang2016linking,finger2016modeling} to integrate multimodal brain networks by constructing a projection between them using GNNs (e.g.,~\cite{zhang2020deep}). 
Most existing studies \cite{zhang2020deep,huang2016linking,finger2016modeling} aim to reconstruct functional networks from brain structural counterparts. 
However, mapping a static object (structural network) to a dynamic one (functional network) is not optimal, even though the structural network may be better suited to serve as the template. 
Very few studies aim to construct the opposite mapping (i.e., from functional to structural networks); this may be because current GNNs can only encode unsigned graphs (i.e., structural networks in which all edge weights have non-negative values). 
The brain functional network is a signed graph by definition (including entries that denote negative correlations); as such, network edge weights can have negative values which will destroy the information aggregation mechanism in current GNNs. 

To tackle this, we propose a new framework to encode signed graphs, and we apply this framework to project and encode brain functional networks to the corresponding structural networks. 
Our results demonstrate that the extracted latent network representations from this mapping can be used for clinical tasks (regression or classification) with better performance, compared with baseline methods. 
Our contributions are three-fold: 
(1) We propose an end-to-end network representation framework to model brain structural networks based on functional ones; this is the first work on this topic to the best of our knowledge. 
(2) We propose a signed graph encoder to embed the functional brain networks. 
(3) We draw graph saliency maps for clinical tasks, to enable phenotypic and disease-related biomarker detection and aid in interpretation.

\section{Preliminaries of Signed Brain Networks}
\label{prelim}
A brain network is an attributed and weighted graph $G=\{V, E\}=(A,X)$ with $N$ nodes, where $V=\{v_i\}_{i=1}^{N}$ is the set of graph nodes representing brain regions, and $E=\{e_{i,j}\}$ is the edge set. 
$X \in \mathcal{R}^{N \times c}$ is the node feature matrix where $x_{i} \in \mathbb{R}^{1 \times c}$ is the $i-$th row of $X$ representing the node feature of $v_{i}$.
$A \in \mathbb{R}^{N \times N}$ is the adjacency matrix where $a_{i,j} \in \mathbb{R}$ represents the weights of the edge between $v_{i}$ and $v_{j}$. 
In signed brain networks (i.e., functional brain networks), $a_{i,j} \in (-\infty, +\infty)$, while in unsigned brain networks (i.e., structural brain networks), $a_{i,j} \in [0, +\infty)$.  
We use $G^{F}=(A^{F}, X^{F})$ and $G^{S}=(A^{S}, X^{S})$ to represent the functional and structural brain networks, respectively. 
Given a specific node $v_{i}$ in a signed network, we define their positive and negative neighbors set as $\mathcal{N}^{+}_{i}$ and $\mathcal{N}^{-}_{i}$, respectively.  
Following the balance theory \cite{cartwright1956structural,derr2018signed,heider1946attitudes,li2020learning}, any node $v_{j}$ belongs to the balanced set (denoted as $\mathbf{\Gamma}$) of $v_{i}$ if the path between $v_{i}$ and $v_{i}$ contains even number of negative edges. 
Otherwise, $v_{j}$ belongs to the unbalanced set (denoted as $\mathbf{\Upsilon}$) of $v_{i}$.

\section{Methodology}
We first introduce the proposed multi-head signed graph encoder (SGE). 
Then, we present an end-to-end framework with the proposed encoder to reconstruct structural networks from functional networks and perform downstream tasks. 
\begin{figure}[t]
\centering
\includegraphics[width=1.0\textwidth]{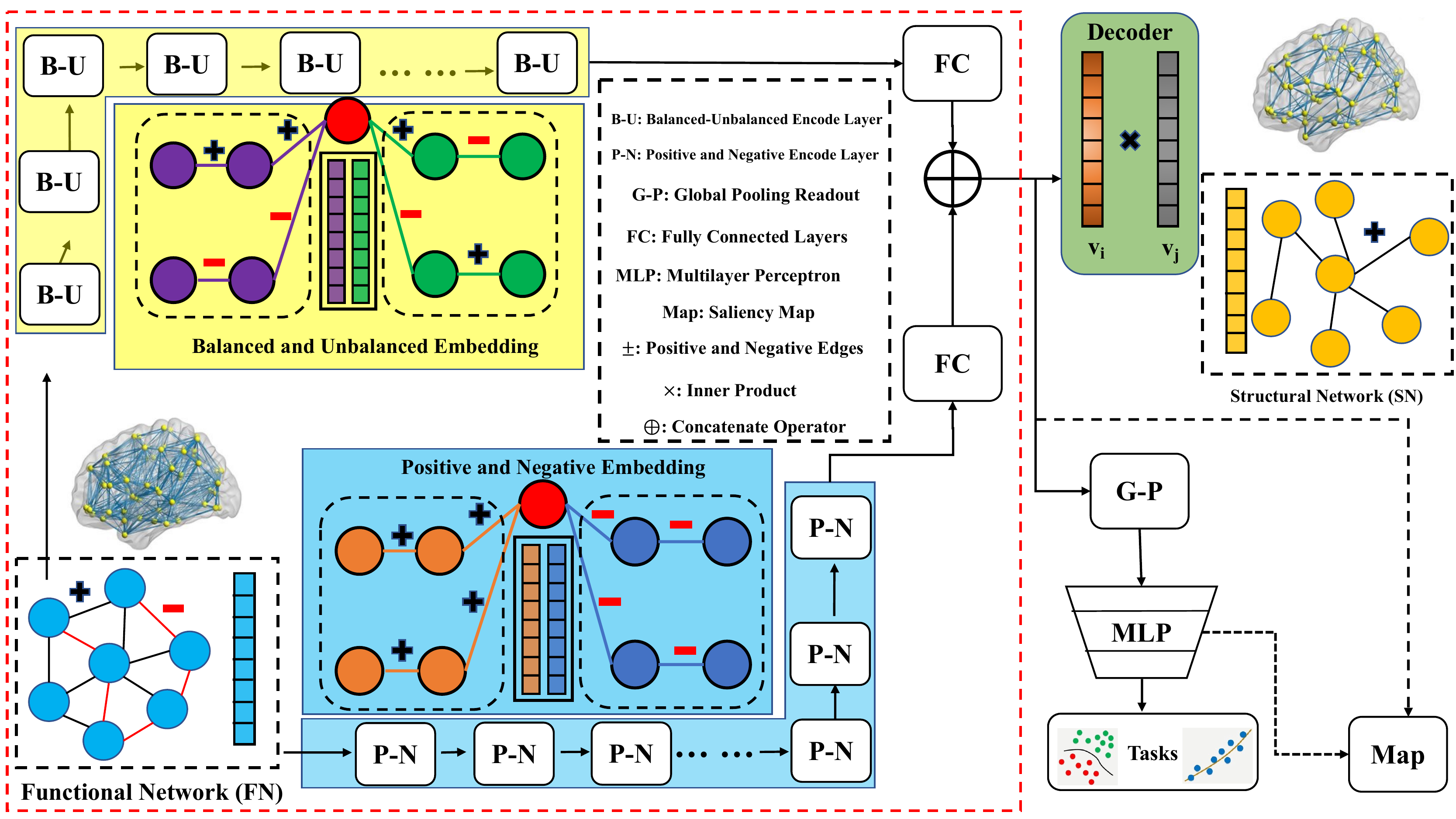}
\caption{Pipeline of the DSBN framework, including a signed graph encoder (in red dash blox) scheme with two branches (BUE and PNE encoder heads in yellow and blue box) for functional network embedding, an inner-product decoder for structural network reconstruction and a downstream task branch for classification and regression.}
\label{framework}
\end{figure}
\subsection{Signed Graph Encoder (SGE)}
Our SGE consists of a balanced-unbalanced encoder (BUE) head and a positive-negative encoder (PNE) head. 
Motivated by the balance theory in Section \ref{prelim}, the BUE head encodes the graph node to the latent features with balanced and unbalanced components.
Meanwhile, we split the signed graph into a positive sub-graph and a negative counterpart. 
The PNE head encodes these two sub-graphs to generate the node latent features with positive and negative components. 
The yellow and blue boxes in Fig. \ref{framework} illustrate the BUE and PNE, respectively.

~\\
\noindent\textbf{Balanced-Unbalanced Encoder (BUE).}
We use $T$ to denote the encoder layer number and the $T-th$ layer of the encoder focuses on aggregating the $T-th$ hop neighbors of node $v_{i}$. 
To improve the generalization of local information aggregation in brain networks, a dynamic and fluctuating adjustment of aggregation weights (i.e., brain network edge weights) is arguably more reasonable and has been advocated during graph learning \cite{zhang2020deep}. 
To this end, our BUE is designed based on the concept of graph attention \cite{velivckovic2017graph}.

\noindent\textbf{\textit{Initial Layer ($T=1$)}}  
We initialize balanced and unbalanced node latent feature components by:
\begin{eqnarray}
    x_{i}^{\mathbf{\Gamma}(1)} = \sigma (\sum_{j \in \mathcal{N}_{i}^{+}} \alpha_{i,j}^{\mathbf{\Gamma}(1)} x^{(0)}_{j} W^{\mathbf{\Gamma}(1)}), \quad
    x_{i}^{\mathbf{\Upsilon}(1)} = \sigma (\sum_{j \in \mathcal{N}_{i}^{-}} \alpha_{i,j}^{\mathbf{\Upsilon}(1)} x^{(0)}_{j} W^{\mathbf{\Upsilon}(1)}), 
\end{eqnarray}
where $\sigma$ is a nonlinear activation function. $x^{(0)}=x$ is input node features. 
$W^{\mathbf{\Gamma}(1)}$ and $ W^{\mathbf{\Upsilon}(1)}$ are trainable weights to generate two latent feature components. $\alpha_{i,j}^{\mathbf{\Gamma}(1)}$ and $\alpha_{i,j}^{\mathbf{\Upsilon}(1)}$ are attention scores of brain nodes from the  balanced and unbalanced set respectively. 
Following the work in \cite{velivckovic2017graph}, we first compute the attention coefficients by:
\begin{eqnarray}
    e_{i,j}^{\mathbf{\Gamma}(1)} = a [x^{(0)}_{i} W^{\mathbf{\Gamma}(1)}, x^{(0)}_{j} W^{\mathbf{\Gamma}(1)}],    \quad  
    e_{i,j}^{\mathbf{\Upsilon}(1)} = a [x^{(0)}_{i} W^{\mathbf{\Upsilon}(1)}, x^{(0)}_{j} W^{\mathbf{\Upsilon}(1)}]
\end{eqnarray}
where $a$ is a trainable attentional weight vector and $[,]$ is a concatenation operator. 
Based on the attention coefficients, we derive the attention scores by:
\begin{eqnarray}
    \alpha_{i,j}^{\mathbf{\Gamma}(1)} = \frac{exp(e_{i,j}^{\mathbf{\Gamma}(1)})}{\sum_{n \in \mathcal{N}_{i}^{+}} exp(e_{i,n}^{\mathbf{\Gamma}(1)})}, \quad
    \alpha_{i,j}^{\mathbf{\Upsilon}(1)} = \frac{exp(e_{i,j}^{\mathbf{\Upsilon}(1)})}{\sum_{n \in \mathcal{N}_{i}^{-}} exp(e_{i,n}^{\mathbf{\Upsilon}(1)})}
\end{eqnarray}

\noindent\textbf{\textit{Following Layers ($T>1$)}} 
In the subsequent layers ($T>1$), $x_{i}^{\mathbf{\Gamma}(T)}$ and $x_{i}^{\mathbf{\Upsilon}(T)}$ is generated by:
\begin{eqnarray}
    x_{i}^{\mathbf{\Gamma}(T)} &=& \sigma (\sum_{j \in \mathcal{N}_{i}^{+}, k \in \mathcal{N}_{i}^{-}} \alpha_{i,j}^{\mathbf{\Gamma}(T)} x^{\mathbf{\Gamma}(T-1)}_{j} W^{\mathbf{\Gamma}(T)} + \alpha_{i,k}^{\mathbf{\Gamma}(T-1)} x^{\mathbf{\Upsilon}(T-1)}_{k} W^{\mathbf{\Gamma}(T)}) \nonumber \\
    x_{i}^{\mathbf{\Upsilon}(T)} &=& \sigma (\sum_{j \in \mathcal{N}_{i}^{+}, k \in \mathcal{N}_{i}^{-}} \alpha_{i,j}^{\mathbf{\Upsilon}(T)} x^{\mathbf{\Upsilon}(T-1)}_{j} W^{\mathbf{\Upsilon}(T)} + \alpha_{i,k}^{\mathbf{\Upsilon}(T-1)} x^{\mathbf{\Gamma}(T-1)}_{k} W^{\mathbf{\Upsilon}(T)})
\end{eqnarray}
Similarly, we compute $4$ attention coefficients as 
\begin{eqnarray}
 e_{i,j}^{\mathbf{\Gamma}(T)} &=& a [x^{\mathbf{\Gamma}(T-1)}_{i} W^{\mathbf{\Gamma}(T)}, x^{\mathbf{\Gamma}(T-1)}_{j} W^{\mathbf{\Gamma}(T)}] \nonumber \\
 e_{i,k}^{\mathbf{\Gamma}(T)} &=& a [x^{\mathbf{\Upsilon}(T-1)}_{i} W^{\mathbf{\Gamma}(T)}, x^{\mathbf{\Upsilon}(T-1)}_{k} W^{\mathbf{\Gamma}(T)}] \nonumber \nonumber \\
 e_{i,j}^{\mathbf{\Upsilon}(T)} &=& a [x^{\mathbf{\Upsilon}(T-1)}_{i} W^{\mathbf{\Upsilon}(T)}, x^{\mathbf{\Upsilon}(T-1)}_{j} W^{\mathbf{\Upsilon}(T)}] \nonumber\\
 e_{i,k}^{\mathbf{\Upsilon}(T)} &=& a [x^{\mathbf{\Gamma}(T-1)}_{i} W^{\mathbf{\Upsilon}(T)}, x^{\mathbf{\Gamma}(T-1)}_{k} W^{\mathbf{\Upsilon}(T)}]
\end{eqnarray}
Then the $4$ attention scores, 
$\alpha_{i,j}^{\mathbf{\Gamma}(T)}=exp(e_{i,j}^{\mathbf{\Gamma}(T)}) / \sum_{\mathbf{\Gamma}(T)}$, 
$\alpha_{i,k}^{\mathbf{\Gamma}(T)}=exp(e_{i,k}^{\mathbf{\Gamma}(T)}) \\ / \sum_{\mathbf{\Gamma}(T)}$, 
$\alpha_{i,j}^{\mathbf{\Upsilon}(T)}=exp(e_{i,j}^{\mathbf{\Upsilon}(T)}) / \sum_{\mathbf{\Upsilon}(T)}$ and
$\alpha_{i,k}^{\mathbf{\Upsilon}(T)}=exp(e_{i,k}^{\mathbf{\Upsilon}(T)}) / \sum_{\mathbf{\Upsilon}(T)}$,
where $\sum_{\mathbf{\Gamma}(T)}=\sum_{n \in \mathcal{N}_{i}} exp(e_{i,n}^{\mathbf{\Gamma}(T)})$ and $\sum_{\mathbf{\Upsilon}(T)}=\sum_{n \in \mathcal{N}_{i}} exp(e_{i,n}^{\mathbf{\Upsilon}(T)})$.

After we obtain the two latent components, we concatenate them as the latent features generated by the BUE head by: $x_{i}^{BUE} = [x_{i}^{\mathbf{\Gamma}}, x_{i}^{\mathbf{\Upsilon}}]$.
~\\
\subsubsection{Positive-Negative Encoder (PNE)}
We split the adjacency matrix of the functional network into positive and negative sub-network pairs (i.e., $G^{+F} =(A^{+F}, X)$ and $G^{-F} =(A^{-F}, X)$). 
For each sub-network, we forward it into $T$ graph attention layers \cite{velivckovic2017graph} to generate the positive and negative latent feature components (i.e., $x_{i}^{+}$ and $x_{i}^{-}$). 
As a result, the latent feature generated by the PNE head can be computed by
$x_{i}^{PNE} = [x_{i}^{+}, x_{i}^{-}]$. 
The final fused latent features generated by our SGE can be computed by: $x_{i} = [x_{i}^{BUE}W, x_{i}^{PNE}W]$, where $W$ is a set of trainable parameters to control feature dimensions.
\vspace{-1em}
\subsection{Deep Signed Brain Networks (DSBN)}
Our DSBN framework is illustrated in Fig. \ref{framework}, which includes (1) a signed graph encoder (SGE) to generate the latent node features from brain functional networks, (2) a decoder to reconstruct brain structural networks from latent node features, and (3) Multilayer Perceptrons (MLP) for downstream tasks.
\vspace{-1em}
\subsubsection{Structural Network Reconstruction and Downstream Tasks}
The structural network edges can be reconstructed by an inner-product decoder\cite{kipf2016variational} as: $\hat{a}_{i,j}^{S} = sigmoid(x_{i}^{\top} \cdot x_{j})$, 
where the $\cdot$ is an inner-product operator, $\top$ is vector transpose. 

We use a sum function as a global readout operator to obtain the whole graph representation (i.e., $X_{G} = \sum_{i=1}^{N} x_{i}$).
Then, MLP are used to generate the final classification or regression output (i.e., $\hat{y} = MLP(X_{G})$). 
Moreover, we use parameters in the last MLP layer and node latent features $x_{i}$ to generate the brain network saliency map using the Class Activation Mapping (CAM) approach \cite{pope2019explainability,arslan2018graph} for the classification task.
The brain saliency map identifies the top-K most important brain regions associated with each class. 
~\\

\noindent\textbf{Loss Functions}
We summarize the loss functions here for our framework. \\
\textbf{\textit{Reconstruction Loss}} 
The ground-truth brain structural network is sparse while the reconstructed structural network is fully connected.
To facilitate network reconstruction, in the training stage, we add a small perturbation value ($\delta$) to the edge weights of the ground-truth structural networks (i.e., $\bar{a}^{S}_{i,j} = a^{S}_{i,j} + \delta$) and build up the reconstruction loss as $    \mathcal{L}_{recon} = \frac{1}{|E|} \sum_{i,j} (\hat{a}_{i,j}^{S} - \bar{a}_{i,j}^{S})^{2},$
where $|E|$ is the number of edges. 

\noindent\textbf{\textit{Supervised Loss}} 
We deploy our framework on both regression and classification tasks. 
For the classification task, we use the negative log likelihood loss where $\mathcal{L}_{super} = NLLLoss(\hat{y}, y)$. 
For the regression task, we use $L_{1}$ loss where $\mathcal{L}_{super} = L_{1}Loss(\hat{y}, y)$.
In summary, the overall loss function of the DSBN is $\mathcal{L}_{all} = \eta_{1} \mathcal{L}_{recon} + \eta_{2} \mathcal{L}_{super},$
where $\eta_{1}$, $\eta_{2}$ are loss weights. 

\section{Experiments}
\begin{figure}[t]
\centering
\includegraphics[width=1.0\textwidth]{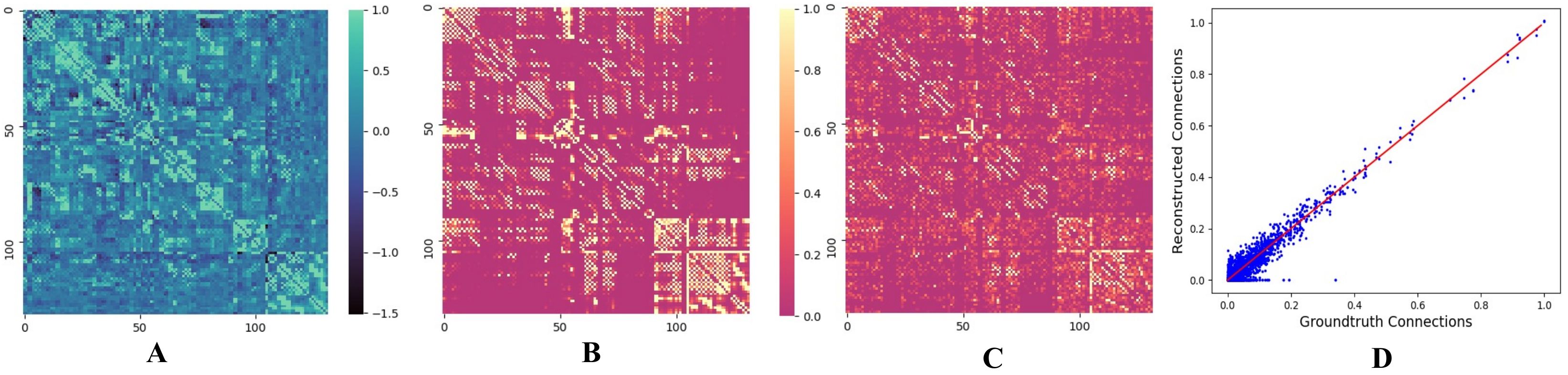}
\caption{Cross-modality learning results on the OASIS data. \textbf{(A)} is the averaged functional network, \textbf{(B)} and \textbf{(C)} are the mean reconstructed and ground-truth structural networks. \textbf{(D)} is to show that edge weights in the predicted structural network are significantly correlated with the ground truth data and $r=0.917$ with $p=0.0129$}
\label{reconstruction}
\end{figure}
\subsection{Data Description and Preprocessing}
Two publicly available datasets were used to evaluate our framework.
The first includes data from $1206$ young healthy subjects (mean age $28.19 \pm 7.15$, $657$ women) from the Human Connectome Project \cite{van2013wu} (HCP). 
The second includes 1326 subjects (mean age $=70.42 \pm 8.95$, $738$ women) from the Open Access Series of Imaging Studies (OASIS) dataset \cite{lamontagne2019oasis}. 
Details of each dataset may be found on their official websites. 
CONN~\cite{whitfield2012conn} and FSL~\cite{jenkinson2012fsl} were used to reconstruct the functional and structural networks, respectively. 
For the HCP data, both networks have a dimension of $82 \times 82$ based on 82 ROIs defined using FreeSurfer (V6.0)~\cite{fischl2012freesurfer}. 
For the OASIS data, both networks have a dimension of $132 \times 132$ based on the Harvard-Oxford Atlas and AAL Atlas. 
We deliberately chose different network resolutions for HCP and OASIS, to evaluate whether the performance of our new framework is affected by the network dimension or atlas. 
\subsection{Implementation Details}
The edge weights of the functional networks and structural networks were first normalized to the range of $[-1, 1]$ and $[0, 1]$, respectively. 
The node features are initialized as the min, $25\%$, median, $75\%$, max values of the fMRI bold signal. 
We randomly split each dataset into $5$ disjoint sets for $5$-fold cross-validations in the following experiments.
The model is trained using the Adam optimizer with a batch size of $128$. 
The initial learning rate is set to $0.001$ and decayed by $(1-\frac{current \, epoch}{max \, epoch})^{0.9}$.
We also regularize the training with an $L_{2}$ weight decay of $1e^{-5}$.
We stop the training if the validation loss does not improve for $100$ epochs in an epoch termination condition with a maximum of $500$ epochs, as was done in \cite{lee2019self,shchur2018pitfalls}.
The experiments are deployed on one NVIDIA TITAN RTX GPU. 
\subsection{Brain Structural Network Reconstruction using DSBN}
To show the performance of our DSBN on the reconstruction of brain structural networks, we train the model in a task-free manner where no task-specific supervised loss is involved. 
Since the $\delta$ value is set to $0.05$ when we train the model, we set the reconstructed edge weights to $0$ if the predicted weights are less than $0.05$.
The mean absolute error (MAE) values between the edge weights in the ground-truth and reconstructed networks are $0.074 \pm 0.016$ and $0.039 \pm 0.058$ under $5$-fold cross-validation on the OASIS and HCP data, respectively. 
The reconstruction results on OASIS data are visualized in Fig. \ref{reconstruction}. 

\begin{table}[t]
\centering
\caption{Classification accuracy, precision, and F1-scores, with their standard deviation values under 5-fold cross-validation. The best results are highlighted in \textbf{bold} font.}
\label{classification}
\resizebox{\textwidth}{21mm}{
\begin{tabular}{l|lll|lll}
\hline
\multicolumn{1}{c|}{Method} & \multicolumn{3}{c|}{OASIS (Disease)}                                                            & \multicolumn{3}{c}{HCP (Gender)}                                                             \\ \hline
                            & \multicolumn{1}{c|}{Acc.}         & \multicolumn{1}{c|}{Prec.}        & \multicolumn{1}{c|}{F1} & \multicolumn{1}{c|}{Acc.}        & \multicolumn{1}{c|}{Prec.}       & \multicolumn{1}{c}{F1} \\ \hline
t-BNE                       & \multicolumn{1}{l|}{$57.84\pm2.14$} & \multicolumn{1}{l|}{$52.26\pm2.39$} & $55.17\pm3.15$ & \multicolumn{1}{l|}{$59.84\pm2.05$} & \multicolumn{1}{l|}{$57.77\pm2.21$} & $55.89\pm1.96$ \\
MK-SVM                      & \multicolumn{1}{l|}{$54.62\pm3.33$} & \multicolumn{1}{l|}{$49.86\pm3.11$} & $55.27\pm4.15$ & \multicolumn{1}{l|}{$58.21\pm3.75$} & \multicolumn{1}{l|}{$51.51\pm2.84$} & $59.19\pm2.40$  \\ \hline
DIFFPOOL                    & \multicolumn{1}{l|}{$66.29\pm2.60$} & \multicolumn{1}{l|}{$63.87\pm2.14$} & $69.91\pm2.37$ & \multicolumn{1}{l|}{$72.12\pm1.81$} & \multicolumn{1}{l|}{$68.84\pm1.97$} & $74.01\pm2.16$ \\
SAGPOOL                     & \multicolumn{1}{l|}{$64.53\pm2.06$} & \multicolumn{1}{l|}{$61.76\pm2.99$} & $68.97\pm3.07$ & \multicolumn{1}{l|}{$69.29\pm1.84$} & \multicolumn{1}{l|}{$70.18\pm1.49$} & $67.38\pm1.12$  \\
BrainChey                   & \multicolumn{1}{l|}{$69.26\pm1.47$} & \multicolumn{1}{l|}{$71.22\pm1.95$} & $70.46\pm2.81$ & \multicolumn{1}{l|}{$74.11\pm2.05$} & \multicolumn{1}{l|}{$76.26\pm1.77$} & $75.08\pm2.60$   \\
BrainNet-CNN                & \multicolumn{1}{l|}{$73.37\pm2.14$} & \multicolumn{1}{l|}{$74.06\pm1.59$} & $73.27\pm1.85$ & \multicolumn{1}{l|}{$71.88\pm1.69$} & \multicolumn{1}{l|}{$70.18\pm2.21$} & $70.29\pm2.10$   \\ \hline
DSBN w/o BUE                & \multicolumn{1}{l|}{$73.04\pm3.03$} & \multicolumn{1}{l|}{$74.74\pm1.96$} & $73.53\pm2.23$ & \multicolumn{1}{l|}{$78.15\pm2.96$} & \multicolumn{1}{l|}{$79.01\pm1.68$} & $80.47\pm2.02$   \\
DSBN w/o PNE                & \multicolumn{1}{l|}{$75.38\pm2.52$} & \multicolumn{1}{l|}{$76.26\pm2.96$} & $78.68\pm3.02$ & \multicolumn{1}{l|}{$80.78\pm2.44$} & \multicolumn{1}{l|}{$81.16\pm1.74$} & $82.98\pm2.01$  \\
DSBN w/o Recon.             & \multicolumn{1}{l|}{$75.94\pm2.73$} & \multicolumn{1}{l|}{$77.09\pm2.52$} & $75.46\pm2.17$ & \multicolumn{1}{l|}{$79.12\pm2.06$} & \multicolumn{1}{l|}{$80.55\pm2.18$} & $81.19\pm1.96$  \\
DSBN   & \multicolumn{1}{l|}{\bm{$78.92\pm1.38$}} & \multicolumn{1}{l|}{\bm{$79.81\pm1.41$}} & \bm{$80.22\pm2.25$} & \multicolumn{1}{l|}{\bm{$82.19\pm2.01$}} & \multicolumn{1}{l|}{\bm{$85.35\pm1.99$}} & \bm{$84.71\pm2.37$}  \\ \hline
\end{tabular}}
\end{table}
\begin{table}[t]
\centering
\caption{Regression Mean Absolute Error (MAE) $\pm$ SD (standard deviation) under 5-fold cross-validation. The best results are highlighted in \textbf{bold} font.}
\label{regression}
\begin{tabular}{l|l|l}
\hline
             & OASIS (MMSE)                    & HCP (MMSE)                     \\ \hline
tBNE & \multicolumn{1}{c|}{$2.39 \pm 0.74$} & \multicolumn{1}{c}{$2.17 \pm 0.48$}\\ 
SAGPOOL & \multicolumn{1}{c|}{$1.86 \pm 0.27$} & \multicolumn{1}{c}{$1.55 \pm 0.33$}\\
DIFFPOOL  & \multicolumn{1}{c|}{$1.69 \pm 0.36$} & \multicolumn{1}{c}{$1.63 \pm 0.14$}\\
BrainNet-CNN  & \multicolumn{1}{c|}{$1.40 \pm 0.20$} & \multicolumn{1}{c}{$1.29 \pm 0.06$}\\ 
BrainChey  & \multicolumn{1}{c|}{$1.12 \pm 0.19$} & \multicolumn{1}{c}{$1.14 \pm 0.25$}\\\hline
DSBN w/o BUE & \multicolumn{1}{c|}{$1.17 \pm 0.22$} & \multicolumn{1}{c}{$1.11 \pm 0.17$}\\
DSBN w/o PNE & \multicolumn{1}{c|}{$1.06 \pm 0.24$} & \multicolumn{1}{c}{$0.86 \pm 0.34$}\\
DSBN w/o Recon. & \multicolumn{1}{c|}{$0.97 \pm 0.11$} & \multicolumn{1}{c}{$1.03 \pm 0.19$}\\
DSBN         & \multicolumn{1}{c|}{\bm{$0.87 \pm 0.18$}} & \multicolumn{1}{c}{\bm{$0.69 \pm 0.21$}} \\ \hline
\end{tabular}
\end{table}
\subsection{Disease and Sex Classification Tasks}
\textbf{\textit{Experimental Setup.}} $6$ baselines were used for comparison, including $2$ traditional graph embedding models (t-BNE \cite{cao2017t} and MK-SVM \cite{dyrba2015multimodal}), $2$ deep graph convolution models designed for brain network embedding (BrainChey \cite{ktena2018metric} and BrainNet-CNN \cite{kawahara2017brainnetcnn}), and $2$ hierarchical
graph neural networks with graph pooling strategies (DIFFPOOL \cite{ying2018hierarchical} and SAGPOOL \cite{lee2019self})
As mentioned above, the baseline methods can only embed unsigned graphs. 
Therefore, we convert the functional networks to unsigned graphs by using the absolute values of the edge weights.  
Meanwhile, $3$ variant models of our DSBN, including DSBN w/o BUE encoder, DSBN w/o PNE encoder and DSBN w/o reconstruction decoder, are evaluated for ablation studies. 
The number of BUE and PNE encoder layers is set to $3$. 
We search the loss weights (details in \textcolor{blue}{Supplementary}) $\eta_{1}$ and $\eta_{2}$ in the range of $[0.01, 0.1, 0.5, 1]$ and $[0.1, 1, 5]$ respectively, and determine the loss weights as $\eta_{1} = 0.1$, $\eta_{2} = 1$ for AD classification, $\eta_{1} = 0.1$, $\eta_{2} = 1$ for sex classification. 
The results are reported in terms of classification accuracy, precision and F1-scores, with their \emph{std}.

\noindent\textbf{\textit{Results}}
Classification results for AD on OASIS and sex on HCP are presented in Table \ref{classification}, which shows that our model achieves the best accuracy for both tasks, among all methods. 
For example, in the AD classification, our model outperforms the baselines with at least $7.6\%$, $7.8\%$ and $9.4\%$ increases in accuracy, precision and F1 scores, respectively. 
In general, the deep graph models perform better than the traditional graph embedding methods (\emph{i.e.}, t-BNE and MK-SVM). 
When we abandon the cross-modality learning (i.e., DSBN w/o Recon.), the performance, though comparable to baselines, decreases significantly. 
This shows the effectiveness of the cross-modality learning.
The performance will also decrease when we remove BUE encoder or PNE encoder, which indicates that both balance and polarity properties are important for embedding functional networks.
The brain saliency map is shown in \textcolor{blue}{Supplementary} where we identify $10$ key brain regions associated with AD (from OASIS) and with each sex (in HCP), respectively. 
The salient regions for females are concentrated in frontal regions of the brain while males have the opposite trend, consistent with the finding ~\cite{carlo1999girls} that women are typically less aggressive than men, and, on average, less physically strong. 
For AD, most of the salient regions are located in subcortical structures, as well as the bilateral intracalcarine region, the caudate, and planum polare, which have been implicated as potential AD biomarkers~\cite{amoroso2017brain}.
\subsection{MMSE Regression}
\textbf{\textit{Experimental Setup}}
Mini-Mental State Exam (MMSE) is a quantitative measure of cognitive status in adults. 
In the MMSE regression task, the selected baselines for comparison (except for MK-SVM which is designed only for classification problems) and the structure of our DSBN remain unchanged.  
The loss weights are set to $\eta_{1}=0.5$ and $\eta_{2}=1$ for both datasets. 
Details of hyperparameter analysis are provided in the \textcolor{blue}{Supplementary}. 
The regression results are reported as average Mean Absolute Errors (MAE) with their \emph{std}.

\noindent\textbf{\textit{Results}}
The MMSE regression results on the HCP and OASIS datasetsare summarized in Table \ref{regression}, which also demonstrates that our model outperforms all baselines. 
The regression results also indicate the superiority of the cross-modality learning and the importance of both BUE and PNE encoders. 


\section{Conclusion}
We propose a novel multimodal brain network representation learning framework with a signed functional network encoder. 
The cross-modality network embedding is generated by mapping a functional brain network to its structural counterpart.
The embedded network representations contribute to important clinical prediction tasks and the brain saliency map may assist with disease-related biomarker identification. 
We will explore the bijection between these two networks in the future.

%
%
%
%
\bibliographystyle{splncs04}
\bibliography{reference}
\end{document}